\documentclass[conference]{IEEEtran}
\IEEEoverridecommandlockouts
\usepackage{cite}
\usepackage{amsmath,amssymb,amsfonts}
\usepackage{algorithmic}
\usepackage{graphicx}
\usepackage{textcomp}
\usepackage{adjustbox}
\usepackage{lipsum} 
\usepackage{booktabs}
\usepackage{multirow}
\usepackage{color}
\usepackage[table,xcdraw]{xcolor}
\usepackage{subcaption}
\usepackage{floatrow}
\usepackage{cite}
\usepackage[T1]{fontenc}
\usepackage[numbers]{natbib}
\usepackage{geometry}

\def\BibTeX{{\rm B\kern-.05em{\sc i\kern-.025em b}\kern-.08em
    T\kern-.1667em\lower.7ex\hbox{E}\kern-.125emX}}

\begin{document}
\newgeometry{top=72pt, left=54pt, right=54pt, bottom=54pt}
\floatsetup[table]{capposition=top}
\newfloatcommand{capbtabbox}{table}[][\FBwidth]

\title{OpenNet: Incremental Learning for Autonomous Driving Object Detection with Balanced Loss\\
}

\author{\IEEEauthorblockN{1\textsuperscript{st} Zezhou Wang}
\IEEEauthorblockA{\textit{MoE Engineering Research Center of}\\
\textit{SW/HW Co-design Technology and Application} \\
\textit{East China Normal University}\\
Shanghai, China \\
10205101573@stu.ecnu.edu.cn}
\and
\IEEEauthorblockN{2\textsuperscript{nd} Guitao Cao$^{\ast}$ \thanks{*Corresponding author}}
\IEEEauthorblockA{\textit{MoE Engineering Research Center of} \\
\textit{SW/HW Co-design Technology and Application} \\
\textit{East China Normal University}\\
Shanghai, China \\
gtcao@sei.ecnu.edu.cn
}
\and
\IEEEauthorblockN{3\textsuperscript{rd} Xidong Xi}
\IEEEauthorblockA{\textit{Software Engineering Institute} \\
\textit{East China Normal University}\\
Shanghai, China \\
52265902004@stu.ecnu.edu.cn
}
\and
\IEEEauthorblockN{4\textsuperscript{th} Jiangtao Wang}
\IEEEauthorblockA{\textit{Software Engineering Institute} \\
\textit{East China Normal University}\\
Shanghai, China \\
jtwang@sei.ecnu.edu.cn}
}
\maketitle

\begin{abstract}
Automated driving object detection has always been a challenging task in computer vision 
due to environmental uncertainties. 
These uncertainties 
include significant differences in object sizes and encountering the class unseen.
It may result in poor performance when traditional object detection models 
are directly applied to automated driving  
detection. Because they usually presume fixed categories of common traffic 
participants, such as pedestrians and cars. Worsely, the huge class imbalance between
 common and novel classes further exacerbates performance degradation. 
 To address the issues stated, we propose OpenNet to moderate 
the class imbalance with the Balanced Loss, which is based on 
Cross Entropy Loss. Besides, we adopt an inductive layer based on gradient reshaping to 
fast learn new classes with limited samples during incremental 
learning. To against catastrophic forgetting, we employ normalized feature distillation. 
By the way, we improve multi-scale detection robustness and unknown class recognition 
through FPN and 
energy-based detection, respectively. The Experimental results upon the CODA dataset show
 that the proposed method can obtain better performance than that of the existing methods.
\end{abstract}


\section{INTRODUCTION}
Autonomous driving detection is a challenging task due to 
the complex, uncertain detection environment, which includes significant discrepancies 
in the uncertainty of class, class imbalance, object sizes and other factors.

The computer vision field has established several common object detection models\cite{b17}-\cite{b19}. 
These models generally assume a relatively balanced sample distribution 
(COCO, VOC) among fixed categories. However, the complexity and variability of 
the real world make it necessary for the model to take into account unknown classes, specifically in 
the automated driving field. It is impossible to guarantee a thorough understanding 
of the classes encountered in the inference environment during training. Additionally, 
training datasets for unknown classes are often 
insufficient as the class imbalance between common and novel classes.

To reconcile these discrepancies, we introduce a new approach to enable object 
detection models for better adaptation to Open World Object Detection. Joseph \cite{b1} compared 
Open Set and Open World, introduced the Open World Object Detection settings. 
Moreover, Joseph introduced a novel methodology, called ORE, 
based on contrastive clustering, an unknown-aware proposal network, and energy-based 
detection of unknown classes. ORE use fine-tuning against catastrophic forgetting, 

However, ORE \cite{b1} has some limitations. Unlike general image detection datasets \cite{b14}, 
datasets in the autonomous driving domain \cite{b20} tend to have large class imbalances 
that do not work using traditional training methods, i.e., direct use of cross-entropy 
classification loss. At the same time, ORE does not work satisfactorily just by 
finetuning to combat forgetting, because finetuning does not directly learn the 
knowledge of the previous model. Worse still, since new classes often contain only a 
limited number of samples, the traditional learning approach does not learn new classes 
well with a limited number of samples.

To address these concerns, we propose a novel method called OpenNet, 
which utilizes meta-learning-based incremental 
learning and Balanced Loss to improve incremental learning and mitigate class imbalance. 
In particular, OpenNet uses normalized feature distillation to combat catastrophic 
forgetting. with improved accuracy by reshaping gradient updates and effective 
learning of new classes under imbalanced samples using meta-learning. Additionally, 
the pyramid structure feature extractor improves the robustness of OpenNet against different scales. 
Finally, we achieved state-of-the-art performance on the CODA \cite{b20} automated driving 
object detection dataset.

Our contributions include:
\begin{itemize}
\item We propose a new classification loss (Balanced Loss) to mitigate class imbalance in autonomous driving object detection.
\item We develop a two-stage framework called OpenNet. The model 
uses an inductive fully connected block (IFC) to reshape gradient updating 
through meta-learning for better model generalization 
and learning capabilities on limited samples
\item Additionally, OpenNet adopts the pyramid feature extractor and normalized feature distillation to improve scale robustness and against catastrophic 
forgetting, respectively. The proposed methodology achieves state-of-the-art performance on the CODA dataset.
\end{itemize}

\section{RELATED WORK}
\subsection{Open World Object Detection}
Dhamija \cite{b5}proposed the first 
formal open-set object detection protocol while introducing a new evaluation 
metric (Average Wilderness Impact) to measure the performance of object detection 
models in open-set scenarios.
Joseph \cite{b1} was the first to introduce Open World Object Detection settings based on 
Open Set and Open World settings. Besides, the evaluation metric (Average Wilderness 
Impact) proposed in Dhamija \cite{b5} was used to measure the performance of unknown 
object detection. 
Similarly, Zheng \cite{b3}proposed a two-stage approach to accomplish unknown detection, first 
predicting known and unknown objects using an open-set object detector, 
then using unsupervised representation learning to discover new objects of unknown 
categories.
However, Zhao \cite{b2} revisited Joseph \cite{b1}, introducing five fundamental benchmark principles 
and two fair evaluation protocols, including two new assessment metrics that 
better evaluate OWOD model performance in handling unknown objects and distinguishing 
difficulties. Additionally, the paper proposes an auxiliary Proposal Advisor (PAD) 
to assist in identifying accurate proposals of unknown classes without supervision, 
and a Class-specific Expelling Classifier (CEC) to calibrate the over-confident 
activation boundary of known classes and filter confusing predictions using a 
class-specific expelling function. Saito \cite{b4}proposed a
new data augmentation and training scheme that enhances
open-world instance segmentation and detection performance.

\subsection{Class Imbalance}
The Focal Loss proposed by Lin \cite{b6} tried to address the class imbalance. 
Focal Loss amplifies the sample weight distribution to any shape of 
perturbation so that minority-class samples can be more finely adjusted.
Based on Lin \cite{b6}, Cui \cite{b7} proposed a new loss function called Class-balanced Loss,
which not only handles binary classification problems but also applies 
to multi-class classification problems.
Also, there are a number of data-side approaches that have been applied to solve the
class imbalance problem \cite{b8}-\cite{b9}.

\subsection{Incremental Learning}
Finn \cite{b10} proposed a new incremental learning method called MAML. 
The method trained the model iteratively on 
multiple tasks to learn initial parameters, allowing 
for faster adaptation to new tasks.
Different from Finn \cite{b10}, Nichol \cite{b11} was based on a special variant of 
gradient descent(Reptile), which can effectively learn from multiple 
tasks and achieve better performance on new tasks. As the datasets are getting larger,
applying previous methods on large-scale datasets straightly is undesirable.
Wu \cite{b12} introduced a method called BiC to correct bias against new categories 
in the fully connected layer by estimating bias parameters. It shows its effectiveness 
on \cite{b13}-\cite{b14} etc.

\subsection{Meta-Learning}
Recently, meta-learning has been applied to address k-shot object detection settings. 
Wang \cite{b15} proposed a simple and effective Fine-Tuning-based approach that 
can leverage the foundation of previous object detectors. Besides, gradient optimization-based 
meta-learning algorithms are widely studied. A MAML-based method called iMAML \cite{b16} used
the implicit gradient approach, and the computational and memory burden of the 
internal optimization loop computation is greatly reduced.

\section{METHODOLOGY}
In this section, we propose our method called OpenNet, which is consist of Balanced Loss, 
and meta-learning-based incremental learning with inductive layers.
\subsection{Incremental Object Detector with Balanced Loss}
\begin{figure}[t]
\centerline{\includegraphics[width=\textwidth]{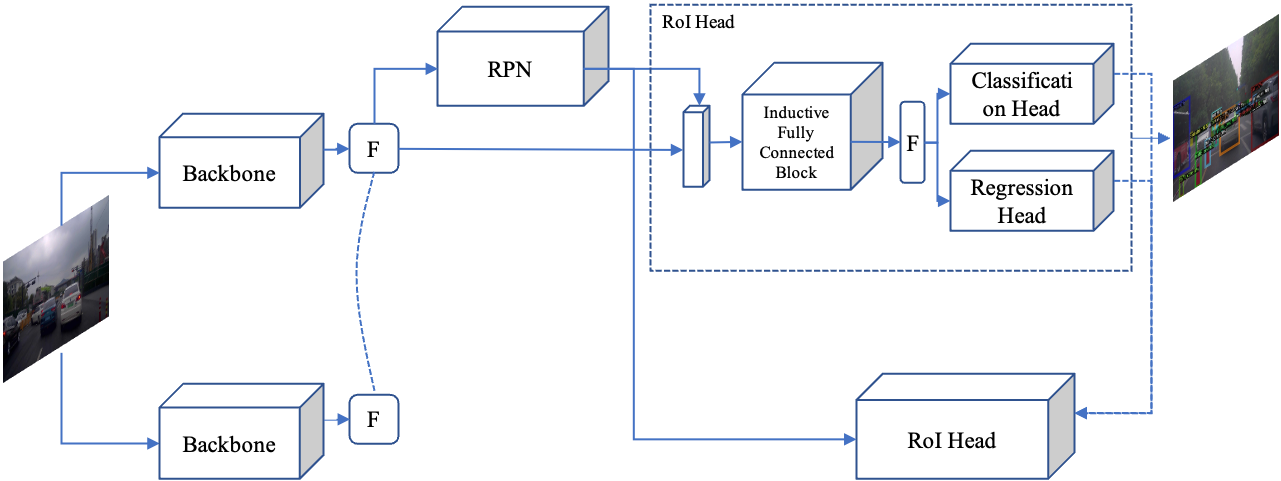}}
\caption {Approach overview. The upper part is the model\textsubscript{{t}} and the 
lower part is the model\textsubscript{{t-1}} with frozen parameters. 
We use model\textsubscript{{t-1}} 
as the teacher for knowledge distillation. The output feature of the Backbone, the 
classification head and the regression head are used as distillation. }  
\label{fig:1}
\end{figure}

Fig.~\ref{fig:1} illustrates the end-to-end structure of incremental object detection.
based on ORE \cite{b1},  OpenNet adopts inductive fully connected blocks (IFC), which accomplish meta-learning 
through inducing gradients. Besides, OpenNet is utilized with multi-scale pyramid feature extractors to enhance 
the robustness against muti-scales. 
The input image is processed by ResNet \cite{b18} 
and FPN \cite{b19} to generate the feature set $ F $, 
which is then separately fed into RPN and RoI Head. RoI Head outputs the class and 
bounding box of each instance respectively. Meanwhile, The output generated by 
RoI Head will be used for contrastive clustering \cite{b1}. Besides, RPN and 
classification head are separately modified to facilitate automatic labeling and 
the ability to recognize unknown classes. After RoI Align, the features are passed to IFC, which
contains two fully connected layers and inductive layers. The inductive layer 
is designed to remodel the gradients. While learning a new task, distilling the 
features outputted by the backbone and the RoI Head of the previous task aids 
against catastrophic forgetting.
We will explain these cohesive components as follows.

\subsection{Balanced Loss}
Lin \cite{b6} found that single-stage detectors have significant accuracy loss compared to 
two-stage detectors and the main reason for this is the huge imbalance between 
foreground and background classes during training. Lin \cite{b6} proposed the 
Focal Loss based on the Cross Entropy Loss to solve the foreground-background class 
imbalance by reducing the weight of the easy-to-classify loss.

The Focal Loss is based on the setting of binary classification task, 
namely foreground and background and the loss function is defined as follows:
\begin{equation}
    Focal\ Loss(p_t)=-\alpha_t(1-p_t)^\gamma log(p_t) \label{eq:1}
\end{equation}
where $ p_t $ is:
\begin{equation}
    p_t = \left\{
            \begin{array}{ll}
                p, & \quad y=1 \\
                1-p, & otherwise \\
            \end{array}
        \right.
    \label{eq:2}
\end{equation}

In order to solve the problem of the imbalance distribution of samples from different 
categories in multi-classification tasks, we propose a new classification Loss called Balanced Loss.

The binary classification tasks often use Sigmoid as the activation function, 
and the output has only one probability value $ p $ of the representative class, then 
the Cross Entropy Loss function based on binary classification is:
\begin{equation}
    Cross\ Entropy\ Loss=-ylog(p)-(1-y)log(1-p) \label{eq:3}
\end{equation}
, which is the loss Focal Loss based on.

The multi-classification tasks, on the other hand, generally use Softmax as the 
activation function and the output has multiple values, which are the probability of 
each class, and the sum of all probability is 1. The Cross Entropy Loss function based on 
multi-classification is then:
\begin{equation}
    Cross\ Entropy\ Loss=-\sum^{C-1}_0 y_i log(p_i) = -log(p_c) \label{eq:4}
\end{equation}
where $ p_c $ is the probability of the correct class.
Thus simply applying Focal Loss to multi-classification is:
\begin{equation}
    Focal\ Loss_{softmax}=-\alpha_c (1-p_c)^\gamma log(p_c) \label{eq:5}
\end{equation}
We note that simply using the transformed Focal Loss \cite{b6} as the loss for the 
multi-classification task can provide a good performance improvement compared to 
the standard Cross Entropy Loss. However, Focal Loss \cite{b6} only serves to reduce the weight of 
the easy-to-classify samples so that their contribution to backpropagation tends 
to 0. Focal Loss \cite{b6} is proposed for single-stage detectors, the ratio 
of front-to-back scenes is often 1:1000 \cite{b6}, yet the ratio in the face of novel and 
common categories are often 1:10,000, which is even more extreme, while Focal Loss for soft examples (easy to classify)
 generally reduces the Loss by only 100 times, which is still less useful in the face 
 of extreme imbalance between the novel and common categories.

As a result, we propose a new categorization loss called Balance Loss as follows:
\begin{equation}
    Balanced\ Loss = - \alpha_c (1-p_c)^\gamma log(p_c)(n-n_c/n) \label{eq:6}
\end{equation}

where $ n $ is the total number of samples, $ n_c $ is the number of samples in the 
correct category, and $ \alpha_c $ is the weight factor. $ (n - n_c/n) $ 
is the balanced factor. We assume that the sample distributions of different categories 
in the dataset simulate the sample distributions of different categories in the real 
world, and even though there are deviations, the deviations are not greater than one 
order of magnitude.

We compared the direct use of sample size for balanced factor calculation and found that 
using the number of instances would be more accurate because a sample may contain 
both a small number of novel classes and a large number of common classes, and using 
the sample size for calculation would hide the true degree of imbalance.

Focal Loss only reduces the contribution of easy-to-categorize categories to the 
loss, while the categorical loss with the addition of Balanced Loss reduces the 
contribution of easy-to-categorize samples to the loss while correspondingly 
increasing the contribution of hard-to-categorize samples to the loss. Compared 
to assigning the same moderator to all categories \cite{b6}, Balanced Loss is more 
flexible in terms of the contribution of different categories to the gradient. 
In summary, Balanced Loss can better cope with extreme sample imbalance compared 
to Focal Loss.

\subsection{Incremental Learning based on Meta-Learning}
In Open World Object Detection, the training set cannot contain samples of all 
possible classes encountered, so incremental learning for new classes needs to 
be performed efficiently to cope with the complex and changing environment 
in the Open World setting \cite{b1}.

However, incremental learning often suffers from catastrophic forgetting, which 
points to a general problem of current target detection models, i.e., the 
stability-plasticity dilemma. In order to solve the stability-plasticity 
dilemma, we need to mitigate catastrophic forgetting on the one hand and use limited 
samples of new classes for efficient incremental learning on the other hand.

Thus, we propose a new module called the inductive fully connected 
block (IFC) to perform meta-learning-based gradient reshaping and against catastrophic 
forgetting through knowledge distillation.

\subsubsection{Task Formulation}
We assume that a standard backpropagation on an object detector $ F $ parameterized by 
$ \theta $ is given by: $ \theta' \leftarrow \theta -\mu\nabla L(\theta) $, where $ L $ 
is the loss function and $ \mu $ is the learning rate. We intend to meta-learn a 
parameterized preconditioning matrix $ I(\theta;\phi) $, where $ \phi $ are parameters 
of I. Therefore, the parameter of $ F $ is set into two sets, which are task parameters
$ \psi $ and inductive parameters $ \phi $. Through the added inductive parameters, the 
gradient pre-conditioning is inherently inductived by the inductive layers.

\begin{figure}[t]
\centerline{\includegraphics[width=\textwidth]{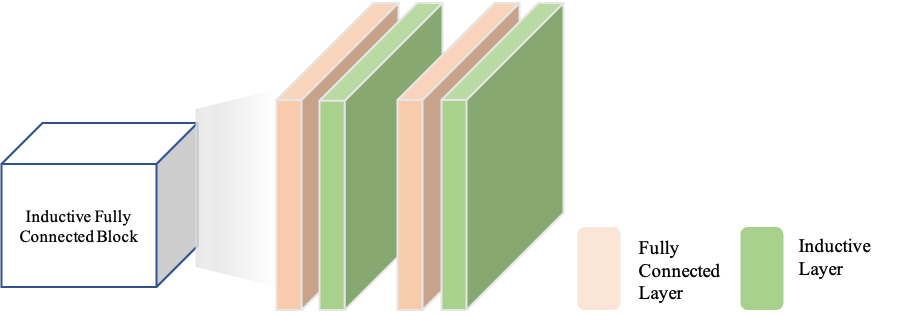}}
\caption {Inductive Layer}  
\label{fig:2}
\end{figure}
We specify some fixed neural network layers as the inductive 
layer, as shown in Fig.~\ref{fig:2}. Because 
the inductive layer is nonlinear, it allows modeling the rich information learned 
by the model in previous training and efficiently reshaping the gradients. It has 
a stronger representation capability than previous work using only diagonal matrices.

The inductive parameters used to preprocess the gradients are trained through all 
tasks of previous training and the goal of reshaping the gradients is achieved 
through our defined inductive loss. These inductive layers help the model to model 
the joint distribution of all the encountered classes, thus achieving 
a better generalization to new tasks and inexplicitly mitigate catastrophic forgetting.

\subsubsection{Task Loss}
We divide Task Loss into two parts, corresponding to the learning of a new task 
and the inheritance of the previous task, which are $ L_{new\_task} $ and $ L_{inherit} $ respectively.

$ L_{new\_task} $ is defined as:
\begin{equation}
    L_{new\_task}=L_{cls}(p,p^*)+\lambda_1L_{loc}(l,l^*)+\lambda_2L_{cont} \label{eq:7}
\end{equation}
$ p $, $ l $ are the output of classification head and regression head respectively, 
$ p^* $, $ l^* $ are the corresponding ground truth, $ \lambda1 $, $ \lambda2$ are the weight 
factors used to balance the various types of losses. Their values are set to 1 
in the experiments. Specifically, $ L_{cls} $ is the Balanced Loss (BL), other losses 
are the same as Joseph \cite{b1}.

For resisting catastrophic forgetting, we use $ model_{t-1} $ with frozen parameters as the teacher 
for knowledge distillation. We choose the output feature of Backbone,  
classification head and regression head as distillation. The input 
images go through $ Backbone_t $ and $ Backbone_{t-1} $ to get $ F_t $ and $ F_t-1 $
 respectively. 
After ROI Heads of $ model_{t} $ and $ model_{t-1} $, we gets $ {p_t,l_t} $ and 
$ {p_{t-1},l_{t-1}} $ respectively. 
As a result, $ L_{inherit} $ is defined as follows:
\begin{equation}
    L_{inherit}=L_{nfd}(F_t,F_{t-1})+L_{KL}(p_t,p_{t-1})+L_{Reg}(l_{t},l_{t-1}) \label{eq:8}
\end{equation}
where  
\begin{equation}
    L_{nfd} = D(Norm(F_t),Norm(F_{t-1})) \label{eq:9}
\end{equation} 
and
\begin{equation}
    Norm(F)=\frac{1}{\sigma}(F-\mu) \label{eq:10}
\end{equation} 
Here, $ F $ is the original feature, $ Norm(F) $ is the normalized feature, $ L_{nfd} $ is the 
normalized feature distillation loss. And 
$ \mu $, $ \sigma $ is the mean and standard deviation of the feature and the default 
is to normalize the feature along the (H, W) dimension. $ D $ is the L2 distance.

$L_{Reg}$ is the  L2  regression loss, and $L_{KL}$ is the KL scatter for computing 
the probability distribution of the current classification head and the output of 
the previous classification head, only for classes that have been previously 
encountered.

The final $ L_{task} $ is a weighted linear combination of $ L_{new\_task} $ and 
$ L_{inherit} $ as 
follows
\begin{equation}
    L_{task}=L_{inherit}+\lambda(L_{new\_task}) \label{eq:11}
\end{equation} 
, $ \lambda $ is the weighting factor to 
balance the distillation loss and the loss of learning the current task. 
We take $ \lambda $ to be 1 in the following experiments.

\subsubsection{Inductive Loss}
For each class $c$ we maintain a queue ($ Base_{c} $), which stores $ N_c $ of $ \{ p,l,p^*,l^* \} $ 
for the computation of Inductive Loss. $ Base_{c} $'s length is $ N_c $. As 
samples are input, $ Base_{c} $ is dynamically updated to maintain the nearest $ N_c $ of 
$ \{ p,l,p*,l* \} $ for each class. Compared to storing a fixed number of samples 
corresponding to each class, this approach ensures that each class contributes 
equally to the update of the induced layer, as common classes and a 
small number of novel classes may appear in the picture, which exacerbates the imbalance. 
Also, because $ Base_{c} $ stores 
the predictions and corresponding annotations of all classes previously encountered. 
It implicitly embeds rich information about classes encountered in previous tasks into 
the inductive layers and the whole network, which also effectively combating 
catastrophic forgetting.

we define $ L_{inductive} $ as follows:
\begin{equation}
    L_{inductive}=\sum_{(p,l,p^*,l^*)\sim Base}L_{cls}(p,p^*)+L_{loc}(l,l^*) \label{eq:12}
\end{equation} 
where $ L_{cls} $ is BL, $ L_{loc} $ is the smooth L1 regression loss.

We want to emphasize that the induced layers and the task layers are updated 
alternately, the induced layers are updated at a fixed interval $iter$, 
and the parameters of the task layers are frozen when the inductive 
layers are updated, and the parameters of the induced layers are 
frozen when the task layer is updated.

\section{EXPERIMENTS AND RESULTS}
We evaluate the proposed approach in the Open World Object Detection 
settings \cite{b1} across two prominent datasets, namely, CODA \cite{b20} 
and SODA10M \cite{b21}. We compare our approach with the state-of-the-art 
method and the standard object detection method, which are ORE \cite{b1} and 
Faster R-CNN \cite{b17} respectively. The result consistently outperforms
 the state-of-the-art method and the standard object detection method. 
Below, we introduce the datasets, explain the experimental settings, 
provide implementation details and report our results. 

\subsection{Datasets}
The experiments were conducted on CODA \cite{b20} and SODA10M \cite{b21}, 
which are two datasets publicly released by Huawei 
to address reliable detection for real-world 
autonomous driving. CODA is divided into CODA Base 
and CODA 2022. CODA Base consists of 1500 camera images, 
while CODA 2022 consists of 
9768 camera images. SODA10M contains 10 million unlabeled images and 
20,000 labeled images. We only used the labeled 
images in SODA10M.

\begin{table}[htbp]
    \caption{EXPERIMENTS DATASETS SPLIT}
    \begin{center}\resizebox{\textwidth}{!}{
    \begin{tabular}{|c|c|c|c|c|c|}
    \hline
        Instances SPLIT & Task1 & Task2 & Task3 & Task4 & Task5 \\ 
    \hline
        train instances & 42377 & 268 & 2290 & 164 & 12 \\ 
    \hline
        test instances & 135 & 341 & 3478 & 413 & 16 \\ 
    \hline
    \end{tabular}
    }
    \label{tab1}
    \end{center}
\end{table}

According to the setup of CODA, SODA10M contains 
six common classes (pedestrian, cyclist, car, 
truck, tram and tricycle), which make up the 
majority of the dataset. The remaining 23 classes 
in CODA 2022, which are in addition to the six 
common classes are considered novel classes.
We group these classes into a set of tasks $ T =
\{ T_1,\ldots T_t,\ldots \} $. All the classes of
 a specific task 
will be introduced to the system at a point in 
time $ t $. While learning $T_t$, all the classes of 
$ {T_\tau: \tau \textgreater t} $ will be treated as unknown.

To achieve the detection of unknown classes and incremental learning 
, we use the data of SODA10M 
as the first task T1. CODA 2022 contains 29 
categories, including a few from the first six classes 
in SODA10M. We group the first six classes, 
which are the common classes in CODA 2022, into 
the first task. The remaining 23 classes are 
divided into groups of 6 classes, except for the last 
group. They are divided into T2 (bus, 
bicycle, moped, motorcycle, stroller and cart), 
T3 (construction vehicle, dog, 
barrier, bollard, sentry box and  
traffic cone), and T4 (traffic island, 
traffic light, traffic sign, debris, 
suitcase, dustbin, concrete block, 
machinery, garbage, plastic bag and 
stone). Because the remaining classes 
contain too few samples to be further 
divided. CODA Base is used as the test set.

As shown in Table \ref{tab1}, the dataset 
is highly imbalanced in terms of common and novel 
classes. This poses a great challenge for traditional
object detection methods.

\subsection{Evaluation Metrics}
\subsubsection{Unknown Classes}
Unknown classes are often mistakenly detected 
as known classes, so we use the Wildness Impact 
(WI) metric \cite{b5} to measure this explicitly. WI is 
defined as:
\begin{equation}
    Wildness \ Impact(WI)=\frac{P_K}{P_{K\cup U}}-1 \label{eq:13}
\end{equation} 
$ P_K $ refers to the accuracy of the model validated on known classes and $ P_{K \cup U} $ refers to 
the accuracy of validation on both known and unknown classes, and when the test set contains 
unknown classes, the lower the WI should be the better.

Also, to measure the absolute numbers of how many unknown classes are classified as known classes, 
we use Absolute Open-Set Error (A-OSE) \cite{b1}.

In summary, we use WI and A-OSE to show the performance of the model in predicting unknown 
classes.

\subsubsection{Known Classes}
The mean average precision at 0.5 IoU threshold (mAP@50) is used as the primary evaluation 
metric for the following experiments.

\subsection{Implementation}
In this paper, OpenNet is pre-trained on the ImageNet dataset \cite{b14} to perform the weight 
initialization of the feature extraction network. We use stochastic gradient descent 
(SGD) with a momentum of 0.9. The initial learning rate was set to 0.01 and subsequently 
reduced to 0.0001 with a warm-up period of 150 iterations. Training is performed on a 
single machine with four GPUs, each processing two images simultaneously, so the effective 
batch size is 8. We set the fixed length of $ Base_{c} $ for each class to 10 and dequeue the 
oldest $ \{p,l,p^*,l^* \} $ when the new $ \{ p, l,p^*,l^* \} $ is added to the particular class.

The classification head of RoI Head only processes the classes encountered so far. According 
to other class incremental work \cite{b22},\cite{b23},\cite{b24}, this is achieved by setting 
the logarithm of unseen classes to a very high negative value ($ -10^{10} $). This allows the 
contribution of the unseen classes in the softmax function to be negligible ($ e^{-10^{10}} \to 0 $) 
while calculating the probability of the classes (referred to as p in \eqref{eq:7}).

The 1024-dim feature vectors from the last 
inductive fully connected block in RoI Head 
are used for contrast clustering. The contrast clustering 
loss (defined in \eqref{eq:7}) is added to the 
classification and localization losses and 
optimized together.

The inductive parameter $ \phi $ is kept fixed 
while the task parameter $ \psi $ is updated and 
vice versa. This is achieved by selectively 
zeroing the gradient during the backward pass 
of the corresponding loss functions $ L_{task} $ and 
$ L_{warp} $. We set $ \alpha $ and $ iter $ to 0.3 and 30, respectively.

We change the structure of Box Head (originally Res-5)\cite{b17} into IFC, 
because of the multi-scale feature extractor 
(FPN), Res-5 was moved to the Backbone. IFC is shown in Fig.~\ref{fig:2} 
, which consists of two sets of fully connected 
layers and inductive layers. The inductive layers are implemented by 
fully connected layers, which input feature 
dimension is the same as the output feature dimension.

\subsection{Results}
\begin{table*}[t]
\caption{EXPERIMENTS RESULTS FOR DIFFERENT METHODS}
\centering
\resizebox{\textwidth}{!}{%
\begin{tabular}{|l|ccc|ccccc|ccccc|ccc|lllllll}
\cline{1-17}
Task IDs                                                                             & \multicolumn{3}{c|}{Task 1}                                                                                                                                                                                                                                & \multicolumn{5}{c|}{Task2}                                                                                                                                                                                                                                                                                                                                                                     & \multicolumn{5}{c|}{Task3}                                                                                                                                                                                                                                                                                                                                                                    & \multicolumn{3}{c|}{Task4}                                                                                                                                                                                                                       &  &  &  &  &  &  &  \\ \cline{1-17}
Evaluation                                                                           & \multicolumn{1}{c|}{WI}                 & \multicolumn{1}{c|}{A-OSE}              & mAP                                                                                            & \multicolumn{1}{c|}{WI}                       & \multicolumn{1}{c|}{A-OSE}                  & \multicolumn{3}{c|}{mAP}                                                                                                                                                                                                                         & \multicolumn{1}{c|}{WI}                       & \multicolumn{1}{c|}{A-OSE}                 & \multicolumn{3}{c|}{mAP}                                                                                                                                                                                                                         & \multicolumn{1}{c|}{}                                                                             & \multicolumn{1}{c|}{mAP}                                                                       & \multicolumn{1}{l|}{}                       &  &  &  &  &  &  &  \\ \cline{2-17}
Metrics                                                                              & \multicolumn{1}{l|}{$\ \ \ \ \downarrow $}& \multicolumn{1}{l|}{$\ \ \ \ \downarrow $ }& \multicolumn{1}{l|}{}                                                                   & \multicolumn{1}{l|}{$\ \ \ \downarrow $}     & \multicolumn{1}{l|}{$\ \ \ \downarrow $}    & \multicolumn{1}{l|}{}                                                                             & \multicolumn{1}{l|}{}                                                                          & \multicolumn{1}{l|}{}                       & \multicolumn{1}{l|}{$\ \ \ \downarrow $}      & \multicolumn{1}{l|}{$\ \ \ \downarrow $}   & \multicolumn{1}{l|}{}                                                                             & \multicolumn{1}{l|}{}                                                                          & \multicolumn{1}{l|}{}                       & \multicolumn{1}{l|}{}                                                                             & \multicolumn{1}{l|}{}                                                                          & \multicolumn{1}{l|}{}                       &  &  &  &  &  &  &  \\
\multirow{-3}{*}{}                                                                   & \multicolumn{1}{l|}{\multirow{-2}{*}} & \multicolumn{1}{l|}{\multirow{-2}{*}} & \multicolumn{1}{l|}{\multirow{-2}{*}{\begin{tabular}[c]{@{}l@{}}Current\\ known\end{tabular}}} & \multicolumn{1}{l|}{\multirow{-2}{*}}       & \multicolumn{1}{l|}{\multirow{-2}{*}}     & \multicolumn{1}{l|}{\multirow{-2}{*}{\begin{tabular}[c]{@{}l@{}}Previously\\ known\end{tabular}}} & \multicolumn{1}{l|}{\multirow{-2}{*}{\begin{tabular}[c]{@{}l@{}}Current\\ known\end{tabular}}} & \multicolumn{1}{l|}{\multirow{-2}{*}{Both}} & \multicolumn{1}{l|}{\multirow{-2}{*}}       & \multicolumn{1}{l|}{\multirow{-2}{*}}    & \multicolumn{1}{l|}{\multirow{-2}{*}{\begin{tabular}[c]{@{}l@{}}Previously\\ known\end{tabular}}} & \multicolumn{1}{l|}{\multirow{-2}{*}{\begin{tabular}[c]{@{}l@{}}Current\\ known\end{tabular}}} & \multicolumn{1}{l|}{\multirow{-2}{*}{Both}} & \multicolumn{1}{l|}{\multirow{-2}{*}{\begin{tabular}[c]{@{}l@{}}Previously\\ known\end{tabular}}} & \multicolumn{1}{l|}{\multirow{-2}{*}{\begin{tabular}[c]{@{}l@{}}Current\\ known\end{tabular}}} & \multicolumn{1}{l|}{\multirow{-2}{*}{Both}} &  &  &  &  &  &  &  \\ \cline{1-17}
Faster-RCNN                                                                          & \multicolumn{1}{c|}{0.07234}            & \multicolumn{1}{c|}{1183}               & 48.71                                                                                          & \multicolumn{1}{c|}{0.0372}                   & \multicolumn{1}{c|}{1002}                   & \multicolumn{1}{c|}{3.044}                                                                        & \multicolumn{1}{c|}{22.16}                                                                     & 13.21                                       & \multicolumn{1}{c|}{0.0222}                   & \multicolumn{1}{c|}{97}                    & \multicolumn{1}{c|}{6.77}                                                                         & \multicolumn{1}{c|}{13.13}                                                                     & 8.29                                        & \multicolumn{1}{c|}{2.12}                                                                         & \multicolumn{1}{c|}{12.51}                                                                     & 3.34                                        &  &  &  &  &  &  &  \\ \cline{1-17}
                                                                                        & \multicolumn{3}{l|}{}                                                                                                                                                                                                            & \multicolumn{1}{c|}{}                         & \multicolumn{1}{c|}{}                       & \multicolumn{1}{c|}{}                                                                             & \multicolumn{1}{c|}{}                                                                          &                                             & \multicolumn{1}{c|}{}              & \multicolumn{1}{c|}{}                      & \multicolumn{1}{c|}{}                                                                             & \multicolumn{1}{c|}{}                                                                          &                                             & \multicolumn{1}{c|}{}                                                                             & \multicolumn{1}{c|}{}                                                                          &                                             &  &  &  &  &  &  &  \\
\multirow{-2}{*}{\begin{tabular}[c]{@{}l@{}}Faster-RCNN\\ + Finetuning\end{tabular}} & \multicolumn{3}{l|}{\multirow{-2}{*}{Task 1 no need to finetune}}                                                                                                                                                           & \multicolumn{1}{c|}{\multirow{-2}{*}{0.0381}} & \multicolumn{1}{c|}{\multirow{-2}{*}{1018}} & \multicolumn{1}{c|}{\multirow{-2}{*}{43.24}}                                                      & \multicolumn{1}{c|}{\multirow{-2}{*}{18.38}}                                                   & \multirow{-2}{*}{36.27}                     & \multicolumn{1}{c|}{\multirow{-2}{*}{0.0287}} & \multicolumn{1}{c|}{\multirow{-2}{*}{101}} & \multicolumn{1}{c|}{\multirow{-2}{*}{35.53}}                                                      & \multicolumn{1}{c|}{\multirow{-2}{*}{10.35}}                                                   & \multirow{-2}{*}{25.54}                     & \multicolumn{1}{c|}{\multirow{-2}{*}{21.27}}                                                      & \multicolumn{1}{c|}{\multirow{-2}{*}{11.49}}                                                   & \multirow{-2}{*}{6.44}                      &  &  &  &  &  &  &  \\ \cline{1-17}
ORE                                                                                  & \multicolumn{1}{c|}{0.02216}            & \multicolumn{1}{c|}{730}                & 49.22                                                                                        & \multicolumn{1}{c|}{0.0163}                   & \multicolumn{1}{c|}{694}                    & \multicolumn{1}{c|}{44.13}                                                                        & \multicolumn{1}{c|}{21.47}                                                                     & 37.17                                       & \multicolumn{1}{c|}{0.0095}                   & \multicolumn{1}{c|}{68}                    & \multicolumn{1}{c|}{35.62}                                                                        & \multicolumn{1}{c|}{12.22}                                                                     & 27.43                                       & \multicolumn{1}{c|}{22.35}                                                                        & \multicolumn{1}{c|}{11.88}                                                                     & 7.66                                        &  &  &  &  &  &  &  \\ \cline{1-17}
OpenNet                                                                              & \multicolumn{1}{c|}{\textbf{0.02014}}   & \multicolumn{1}{c|}{\textbf{642}}       & \textbf{50.12}                                                                            & \multicolumn{1}{c|}{\textbf{0.0144}}          & \multicolumn{1}{c|}{\textbf{601}}           & \multicolumn{1}{c|}{\textbf{46.32}}                                                               & \multicolumn{1}{c|}{\textbf{22.72}}                                                            & \textbf{37.94}                              & \multicolumn{1}{c|}{\textbf{0.0088}}          & \multicolumn{1}{c|}{\textbf{55}}           & \multicolumn{1}{c|}{\textbf{38.88}}                                                               & \multicolumn{1}{c|}{\textbf{13.41}}                                                            & \textbf{28.81}                              & \multicolumn{1}{c|}{\textbf{23.54}}                                                               & \multicolumn{1}{c|}{\textbf{12.66}}                                                            & \textbf{8.81}                               &  &  &  &  &  &  &  \\ \cline{1-17}
\end{tabular}%
\label{tab2}
}
\end{table*}

We select ORE \cite{b1}, Faster R-CNN \cite{b17} as BaseLine. 
ORE \cite{b1} currently has achieved the SOTA 
performance in Open World Object Detection setting.
 Besides, Faster R-CNN is selected as the baseline 
 of standard object detector. 

The results of the experiments are shown in 
Table \ref{tab2}. Here we show how different 
methods perform in Open World Object Detection settings. 
The WI and A-OSE measure OpenNet's performance 
on unknown classes. Meanwhile, mAP measures how 
well OpenNet detects both current classes 
and previous classes. 

It's previous that OpenNet 
outperforms both Faster R-CNN \cite{b17} and ORE \cite{b1} on all 
validation metrics, which finally get 8.81\% mAP. Faster R-CNN, Faster R-CNN+Finetuning and ORE get
 3.34\%, 6.44\% and 7.66\% mAP respectively. 
 Besides, We can see that the 
standard target detector (Faster-RCNN) demonstrates 
catastrophic forgetting with task switching in the 
face of Open World Object Detection settings. 
Compared to ORE and OpenNet, there is a huge gap in its ability to detect previous tasks
 (3.044\%, 6.77\% and 2.12\% mAP respectively). 
Since ORE is based on Faster-RCNN sample playback, 
i.e., Finetuning, its ability to detect previous 
classes is close to Faster R-CNN+Finetuning, while OpenNet 
has better performance in 
combating catastrophic forgetting because of the 
knowledge distillation. Also, as the added BL and 
the standard ROI Head is replaced by the ROI Head 
with the ability to reshape gradients, i.e., consisting of IFC, etc. 
It can be seen that 
these enable OpenNet has stronger generalization and 
faster learning capability, i.e., stronger ability 
to learn, in the face of extremely class imbalance. Finally, with the addition 
multi-scale pyramid structure, the 
robustness of multi-scale detection of OpenNet is further enhanced.

\begin{figure}[t]
\centerline{\includegraphics[width=\textwidth]{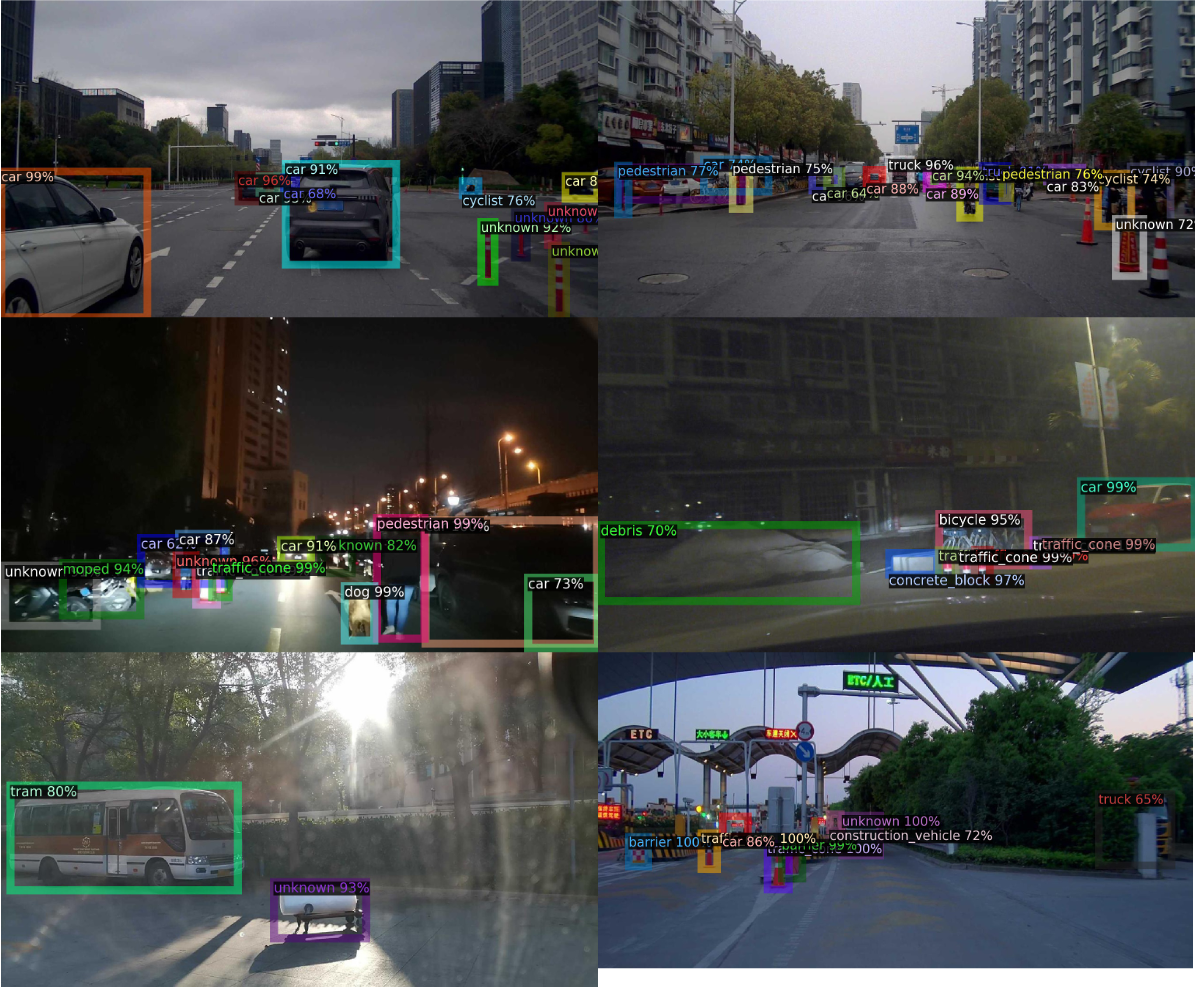}}
\caption{The results of the experiments.}
\label{Figure 3}
\end{figure}

Some experimental results based on the 6+23 
experimental settings at CODA Base, i.e. T1 
for common classes and T2 for novel classes. 
As shown in Fig.~\ref{Figure 3}, the first row shows the results of training on 
common classes only, which shows that the model 
can only recognize common classes at this time, 
and can recognize some novel classes as unknown 
classes. The second and third rows show the 
detection results after the model learned the 
novel class at T2. We can see that OpenNet with 
the multi-scale feature extractor can detect 
the objects very well, no matter if they are large 
or small in size, reflecting good robustness. 
At the same time, it can be seen that with a 
small number of novel classes as the training 
set, it also has a better learning generalization ability 
for novel classes while well retaining 
the learning information from the original common 
classes. In summary, the pictures show the 
effectiveness of Balanced Loss, knowledge 
distillation and inductive layers by visualization.

\section{DISCUSSIONS AND ANALYSIS}
To test the effectiveness of Balanced Loss
 compared to Focal Loss and the effectiveness 
 of adding IFC, we use ablation experiments. The ablation experiments 
 use the 6+23 setting, which is also used in the 
 sensitivity experiments later on. Since the 
 performance comparison here is not related
 to whether the unknown class is detected or 
 not, mAP@50 is used as the evaluation metric. 
 Meanwhile, Standard Cross Entropy Loss will be used as 
 a replacement while Balanced Loss is eliminated.

\subsection{Ablation Experiments}
\begin{table*}\small
\begin{floatrow}
\capbtabbox{
    \begin{tabular}{|l|l|l|l|l|l|}
    \hline
    B   & D   & I   & T1             & T2             & ALL            \\ \hline
    no  & yes & yes & 49.56          & 14.43          & 21.89          \\ \hline
    yes & no  & yes & 49.77          & 15.32          & 16.11          \\ \hline
    yes & yes & no  & 48.92          & 15.21          & 22.16          \\ \hline
    yes & yes & yes & \textbf{50.12} & \textbf{16.61} & \textbf{24.24} \\ \hline
    \end{tabular}
}{
    \caption{THE ABLATION EXPERIMENTS RESULTS ON BALANCED LOSS, DISTILLATION AND INDUCTIVE LAYER}
    \label{tab3}
    \small
}
\capbtabbox{
    \begin{tabular}{|l|l|l|l|}
    \hline
    Methods                                                                                                & T1                     & T2                     & ALL                    \\ \hline
    OpenNet                                                                                         & \textbf{50.12}         & \textbf{16.61}         & \textbf{24.24}         \\ \hline
    \multirow{2}{*}{\begin{tabular}[c]{@{}l@{}}OpenNet\\ +Cross Entropy Loss\end{tabular}}          & \multirow{2}{*}{49.33} & \multirow{2}{*}{15.12} & \multirow{2}{*}{21.13} \\
                                                                                                    &                        &                        &                        \\ \hline
    \multirow{2}{*}{\begin{tabular}[c]{@{}l@{}}OpenNet\\ +Weighted Cross Entropy Loss\end{tabular}} & \multirow{2}{*}{49.45} & \multirow{2}{*}{15.34} & \multirow{2}{*}{22.11} \\
                                                                                                    &                        &                        &                        \\ \hline
    \multirow{2}{*}{\begin{tabular}[c]{@{}l@{}}OpenNet\\ +FocalLoss\end{tabular}}                   & \multirow{2}{*}{49.89} & \multirow{2}{*}{15.67} & \multirow{2}{*}{23.27} \\
                                                                                                    &                        &                        &                        \\ \hline
    \end{tabular}
}{
    \caption{THE ABLATION EXPERIMENTS RESULTS ON INCREMENTAL LEARNING AND CLASSIFICATION LOSS}
    \label{tab4}
}

\end{floatrow}
\end{table*}

We use ablation experiments to understand the contribution of each of the constituent 
components of the proposed methodology: Balanced Loss, knowledge distillation and IFC.
 Table \ref{tab3} shows the results of the ablation 
experiments, where whether the constituent components are selected or not is indicated 
by yes and no, and B, D and I refer to Balanced Loss, knowledge distillation and IFC, 
respectively.

We see that a mAP of 22.89\% can be achieved with the elimination of Balanced Loss, i.e.,
 using the standard cross-entropy, and 24.24\% with the inclusion of Balanced Loss, 
 which is a 1.35\% improvement. While a mAP of 22.16\% can be achieved without adding 
 the induced layer, a higher mAP of 24.24\% can be achieved with the addition of the induced 
 layer, i.e., there is a 2.08\% improvement. This reflects the more efficient generalization 
 and learning ability of the model using Balanced Loss and IFC in the face 
 of class imbalance.

Ultimately, although using both Balanced Loss and IFC in both T1 and T2 tasks 
compared to removing either one can achieve a higher mAP of 49.77\% and 15.32\%, 
respectively, the process of incremental learning eventually leads to a lower mAP of 
16.11\% due to catastrophic forgetting. This exemplifies the effectiveness of knowledge 
distillation in combating catastrophic forgetting.

To further demonstrate the effectiveness of Balanced Loss, we conducted ablation 
experiments for classification loss, while other settings remain unchanged. 
Here we use OpenNet with standard Cross Entropy Loss as Baseline. As shown in Table \ref{tab4}, 
there is a 0.98\% improvement compared with Baseline using weighted 
Cross Entropy Loss. But it is still not ideal. Focal Loss is from the perspective of 
easy and difficult samples. Lin \cite{b6} concludes that most of the loss comes from simple 
negative samples and the gradient direction is dominated by these samples. Focal Loss get a 2.14\% 
improvement compared to Baseline. Balanced Loss is a combination of the common and novel 
classes and the hard and easy samples. BL increases the weight of the hard samples 
and decreases the weight of the easy samples while increasing the weight of the novel 
classes. Balanced Loss has 3.11\%, 2.13\% and 0.97\% improvement over Standard Cross 
Entropy Loss, Weighted Cross Entropy Loss and Focal Loss, respectively, which fully 
demonstrates the superiority of Balanced Loss in the extreme class imbalance setting.

\subsection{Sensitivity Analysis}
\begin{table*}[]\small
\begin{floatrow}
\capbtabbox{
\begin{tabular}{|c|c|c|c|c|c|c|c|c|c|}
\hline
        &\multicolumn{3}{c|}{$iter$=30}                & \multicolumn{3}{c|}{$iter$=300}       & \multicolumn{3}{c|}{$iter$=3000}       \\ \hline
$\alpha$ & T1       & T2    & Both           & T1         & T2    & Both  & T1          & T2    & Both  \\ \hline
0.1    & 49.77    & 16.41 & 24.21          & 49.68      & 16.21 & 24.17 & 48.57       & 16.02 & 23.99 \\ \hline
0.2    & 49.89    & 16.44 & 24.44          & 48.87      & 16.37 & 24.09 & 48.74       & 16.23 & 23.71 \\ \hline
0.3    & 50.12    & 16.61 & \textbf{25.12} & 49.11      & 16.51 & 24.19 & 48.97       & 16.48 & 23.97 \\ \hline
0.4    & 49.24    & 16.56 & 24.18          & 48.28      & 15.84 & 23.87 & 48.19       & 15.78 & 23.46 \\ \hline
0.6    & 48.94    & 15.97 & 23.87          & 48.57      & 15.76 & 23.55 & 48.47       & 15.71 & 23.14 \\ \hline
0.8    & 48.88    & 15.88 & 23.55          & 48.45      & 15.69 & 23.21 & 48.23       & 15.65 & 22.94 \\ \hline
\end{tabular}
}{
\caption{SENSITIVITY ANALYSIS ON HYPER-PARAMETERS $\alpha$ AND $iter$}
\label{tab5}
}
\end{floatrow}
\end{table*}
We performed sensitivity analysis at the weighting factor $\alpha$ 
which weighs the importance of the distillation loss $L_{inherit}$ and $L_{new\_task}$ in 
\eqref{eq:11}. and the update interval $iter$ of $ L_{inductive} $, all experiments were performed 
in the same experimental setting of 6+23 as the ablation 
experiments. The results of the sensitivity analysis are presented in Table \ref{tab5}.

In the vertical analysis, with a constant update frequency of the inductive layer, 
Table \ref{tab5} shows that the model's ability to learn new tasks tends to decrease as 
the importance of knowledge distillation increases, which is consistent with iter 
= 30/300/3000. In the cross-sectional analysis, with constant knowledge distillation 
weighting coefficients, the learning performance of the model in new classes gradually 
decreases as the frequency of updating the inductive layers decreases, and it is clear 
that more frequent inductive layer updates can induce gradients more effectively and 
enable the model to learn the optimal solution from prvious tasks.

\section{CONCLUSION}
In the Open World Object Detection setting, ORE achieves the recognition of unknown classes as well as 
incremental learning. In this work, in order to solve the class imbalance in 
self-driving object detection, we first propose Focal Loss to alleviate the class 
imbalance by reducing the weight of common classes and soft samples 
while increasing the weight of novel classes and hard samples from the 
perspective of common classes and novel classes as well as hard and easy samples. Meanwhile, 
for more effective incremental learning, we propose incremental learning based on 
meta-learning, which reshapes the gradient by adding the inductive fully connected block, allowing 
information to be shared implicitly across incremental tasks. Together with knowledge 
distillation, this allows the model to not only retain prior knowledge well, but also 
learn new classes with only finite new class samples. In addition, our extensive comparison experiments, 
ablation experiments and sensitivity experiments show the independent contribution of 
each component of the method. Extending our method to single-stage detectors\cite{b26}-\cite{b28} as well as 
other computer vision tasks such as instance segmentation\cite{b25} would be interesting and important 
research directions.

\section{ACKNOWLEDGMENTS}
This work was supported by the National Natural Science Foundation of China under Grant 61871186 and 61771322.

\vspace{12pt}


\begin{thebibliography}{00}
\bibitem{b1} Joseph, K. J., et al. "Towards open world object detection." Proceedings of the IEEE/CVF Conference on Computer Vision and Pattern Recognition. 2021.
\bibitem{b2} Zhao, Xiaowei, et al. "Revisiting open world object detection." arXiv preprint arXiv:2201.00471 (2022).
\bibitem{b3} Zheng, Jiyang, et al. "Towards open-set object detection and discovery." Proceedings of the IEEE/CVF Conference on Computer Vision and Pattern Recognition. 2022.
\bibitem{b4} Saito, Kuniaki, et al. "Learning to detect every thing in an open world." Computer Vision–ECCV 2022: 17th European Conference, Tel Aviv, Israel, October 23–27, 2022, Proceedings, Part XXIV. Cham: Springer Nature Switzerland, 2022.
\bibitem{b5} Dhamija, Akshay, et al. "The overlooked elephant of object detection: Open set." Proceedings of the IEEE/CVF Winter Conference on Applications of Computer Vision. 2020.

\bibitem{b6} Lin, Tsung-Yi, et al. "Focal loss for dense object detection." Proceedings of the IEEE international conference on computer vision. 2017.
\bibitem{b7} Cui, Yin, et al. "Class-balanced loss based on effective number of samples." Proceedings of the IEEE/CVF conference on computer vision and pattern recognition. 2019.
\bibitem{b8} Fernández, Alberto, et al. "SMOTE for learning from imbalanced data: progress and challenges, marking the 15-year anniversary." Journal of artificial intelligence research 61 (2018): 863-905.
\bibitem{b9} Liu, Tian-Yu. "Easyensemble and feature selection for imbalance data sets." 2009 international joint conference on bioinformatics, systems biology and intelligent computing. IEEE, 2009.

\bibitem{b10} Finn, Chelsea, Pieter Abbeel, and Sergey Levine. "Model-agnostic meta-learning for fast adaptation of deep networks." International conference on machine learning. PMLR, 2017.
\bibitem{b11} Nichol, Alex, Joshua Achiam, and John Schulman. "On first-order meta-learning algorithms." arXiv preprint arXiv:1803.02999 (2018).
\bibitem{b12} Wu, Yue, et al. "Large scale incremental learning." Proceedings of the IEEE/CVF Conference on Computer Vision and Pattern Recognition. 2019.
\bibitem{b13} Krizhevsky, Alex, and Geoffrey Hinton. "Learning multiple layers of features from tiny images." (2009): 7.
\bibitem{b14} Deng, Jia, et al. "Imagenet: A large-scale hierarchical image database." 2009 IEEE conference on computer vision and pattern recognition. Ieee, 2009.
\bibitem{b15} Wang, Xin, et al. "Frustratingly simple few-shot object detection." arXiv preprint arXiv:2003.06957 (2020).
\bibitem{b16} Rajeswaran, Aravind, et al. "Meta-learning with implicit gradients." Advances in neural information processing systems 32 (2019).

\bibitem{b17} Ren, Shaoqing, et al. "Faster r-cnn: Towards real-time object detection with region proposal networks." Advances in neural information processing systems 28 (2015).
\bibitem{b18} He, Kaiming, et al. "Deep residual learning for image recognition." Proceedings of the IEEE conference on computer vision and pattern recognition. 2016.
\bibitem{b19} Lin, Tsung-Yi, et al. "Feature pyramid networks for object detection." Proceedings of the IEEE conference on computer vision and pattern recognition. 2017.

\bibitem{b20} Li, Kaican, et al. "Coda: A real-world road corner case dataset for object detection in autonomous driving." Computer Vision–ECCV 2022: 17th European Conference, Tel Aviv, Israel, October 23–27, 2022, Proceedings, Part XXXVIII. Cham: Springer Nature Switzerland, 2022.
\bibitem{b21} Han, Jianhua, et al. "SODA10M: a large-scale 2D self/Semi-supervised object detection dataset for autonomous driving." arXiv preprint arXiv:2106.11118 (2021).

\bibitem{b22} Rajasegaran, Jathushan, et al. "Random path selection for continual learning." Advances in Neural Information Processing Systems 32 (2019).
\bibitem{b23} Lopez-Paz, David, and Marc'Aurelio Ranzato. "Gradient episodic memory for continual learning." Advances in neural information processing systems 30 (2017).
\bibitem{b24} Chaudhry, Arslan, et al. "Efficient lifelong learning with a-gem." arXiv preprint arXiv:1812.00420 (2018).

\bibitem{b25} Cholakkal, Hisham, et al. "Object counting and instance segmentation with image-level supervision." Proceedings of the IEEE/CVF Conference on Computer Vision and Pattern Recognition. 2019.
\bibitem{b26} Nie, Jing, et al. "Enriched feature guided refinement network for object detection." Proceedings of the IEEE/CVF international conference on computer vision. 2019.
\bibitem{b27} Wang, Tiancai, et al. "Learning rich features at high-speed for single-shot object detection." Proceedings of the IEEE/CVF international conference on computer vision. 2019.
\bibitem{b28} Pang, Yanwei, et al. "Efficient featurized image pyramid network for single shot detector." Proceedings of the IEEE/CVF conference on computer vision and pattern recognition. 2019.
\end{thebibliography}
\end{document}